\documentclass{article}

\usepackage{PRIMEarxiv}

\usepackage[utf8]{inputenc} 
\usepackage[T1]{fontenc}    
\usepackage{hyperref}       
\usepackage{url}            
\usepackage{booktabs}       
\usepackage{amsfonts}       
\usepackage{nicefrac}       
\usepackage{microtype}      
\usepackage{lipsum}
\usepackage{fancyhdr}       
\usepackage{graphicx}       
\graphicspath{{media/}}     
\usepackage{svg}
\usepackage{doi}
\usepackage{amsmath}
\usepackage{multicol}

\usepackage{xcolor}

\newcounter{countitems}
\newcounter{nextitemizecount}
\newcommand{\setupcountitems}{%
  \stepcounter{nextitemizecount}%
  \setcounter{countitems}{0}%
  \preto\item{\stepcounter{countitems}}%
}
\makeatletter
\newcommand{\computecountitems}{%
  \edef\@currentlabel{\number\c@countitems}%
  \label{countitems@\number\numexpr\value{nextitemizecount}-1\relax}%
}
\newcommand{\nextitemizecount}{%
  \getrefnumber{countitems@\number\c@nextitemizecount}%
}
\newcommand{\previtemizecount}{%
  \getrefnumber{countitems@\number\numexpr\value{nextitemizecount}-1\relax}%
}
\makeatother    
{\end{itemize}%
\unskip\computecountitems\ifnumcomp{\previtemizecount}{>}{3}{\end{multicols}}{}}

\usepackage{float}
\usepackage{aliascnt}
\newaliascnt{eqfloat}{equation}
\newfloat{eqfloat}{h}{eqflts}
\floatname{eqfloat}{Equation}

\newcommand*{\ORGeqfloat}{}
\let\ORGeqfloat\eqfloat
\def\eqfloat{%
  \let\ORIGINALcaption\caption
  \def\caption{%
    \addtocounter{equation}{-1}%
    \ORIGINALcaption
  }%
  \ORGeqfloat
}
\title{ABANICCO: A New Color Space for Multi-Label Pixel Classification and Color Segmentation
}

\author{ \href{https://orcid.org/0000-0003-4731-8262}{\includegraphics[scale=0.06]{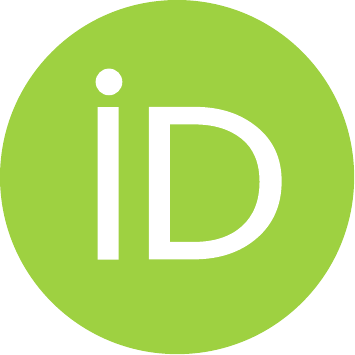}\hspace{1mm}Laura Nicolás-Sáenz}\thanks{Corresponding author} \\
	Departamento de Bioingeniería\\
	Universidad Carlos III de Madrid\\
	Leganés, Spain \\
	\texttt{lnicolas@ing.uc3m.es} \\
	\And
	\href{https://orcid.org/0000-0000-0000-0000}{\includegraphics[scale=0.06]{orcid.pdf}\hspace{1mm}Agapito Ledezma} \\
	Departmento de Informática\\
	Universidad Carlos III de Madrid\\
	Leganés, Spain \\
	\texttt{ledezma@inf.uc3m.es } \\
		\And
	\href{https://orcid.org/0000-0003-1484-731X}{\includegraphics[scale=0.06]{orcid.pdf}\hspace{1mm}Javier Pascau} \\
	Departamento de Bioingeniería\\
	Universidad Carlos III de Madrid\\
	Leganés, Spain \\
	\texttt{jpascau@ing.uc3m.es } \\
		\And
	\href{https://orcid.org/0000-0002-1573-1661}{\includegraphics[scale=0.06]{orcid.pdf}\hspace{1mm}Arrate Muñoz-Barrutia} \\
	Departamento de Bioingeniería\\
	Universidad Carlos III de Madrid\\
	Leganés, Spain \\
	\texttt{mamunozb@ing.uc3m.es } \\
}

\begin{document}
\maketitle

\begin{abstract}
In any computer vision task involving color images, a necessary step is classifying pixels according
to color and segmenting the respective areas. However, the development of methods able to successfully complete this task has proven challenging, mainly due to the gap between human color perception,
linguistic color terms, and digital representation. In this paper, we propose a novel method combining
geometric analysis of color theory, fuzzy color spaces, and multi-label systems for the automatic
classification of pixels according to 12 standard color categories (Green, Yellow, Light Orange, Deep
Orange, Red, Pink, Purple, Ultramarine, Blue, Teal, Brown, and Neutral). Moreover, we present a robust,
unsupervised, unbiased strategy for color naming based on statistics and color theory. ABANICCO was tested against the state of the art in color classification and with the standarized  ISCC–NBS color system, providing accurate classification and  a standard, easily understandable alternative for hue naming recognizable by humans and machines. We expect this solution to become the
base to successfully tackle a myriad of problems in all fields of computer vision, such as region
characterization, histopathology analysis, fire detection, product quality prediction, object description, and hyperspectral imaging.
\end{abstract}


\section{Introduction}
Color classification of image pixels can be used to resolve various tasks in a wide range of fields, from medical histopathology to satellite imaging \cite{ganesan2017user}, including even ripeness prediction of fresh produce \cite{dadwal2012estimate}. Many algorithms have been proposed to this end. However, we still lack one fully-tested automatic method that can be applied to various problems and is not based on individual human perception and supervision. This is due to the intrinsic difficulties of working with colors: the lack of clear boundaries between different hues, the apparent subjectivity of human perception, the cultural and contextual differences, and the absence of consensus regarding color naming. As Newton said, “color is a sensation in the viewers' mind” \cite{grandy2005goethe}.

Color naming as a strategy to close the semantic gap between the language humans and machines apply for color is arduous. It is challenging to map linguistic labels of the multidimensional color space to computational representations, mainly because even humans have trouble agreeing on standard, universal names for said colors. Furthermore, colors are continuous in nature, with blurred boundaries between the possible color categories.

To tackle these problems, we propose a method for pixel classification and color segmentation independent of human interaction based on classical color theory. Starting from this background, accepted by classical artists and modern researchers, we have created a novel method to identify and classify colors in digital images for subsequent segmentation. Thanks to classical color theory, we can carry out these tasks without requiring human labels or supervision that would be affected by subjective perception, providing a final  result as objective as possible. Moreover, we name the identified colors homogeneously and understandably by applying notions from basic mathematics and art. In this way, we can close the semantic gap between humans and machines. Our proposed method is called ABANICCO: an AB ANgular Illustrative Classification of COlor.

To adequately explain our work, we must first introduce the fundamental concepts of classical color theory, color naming, and modern color models.

\begin{figure}[hb!]
  \centering
  \includegraphics[scale=0.5]{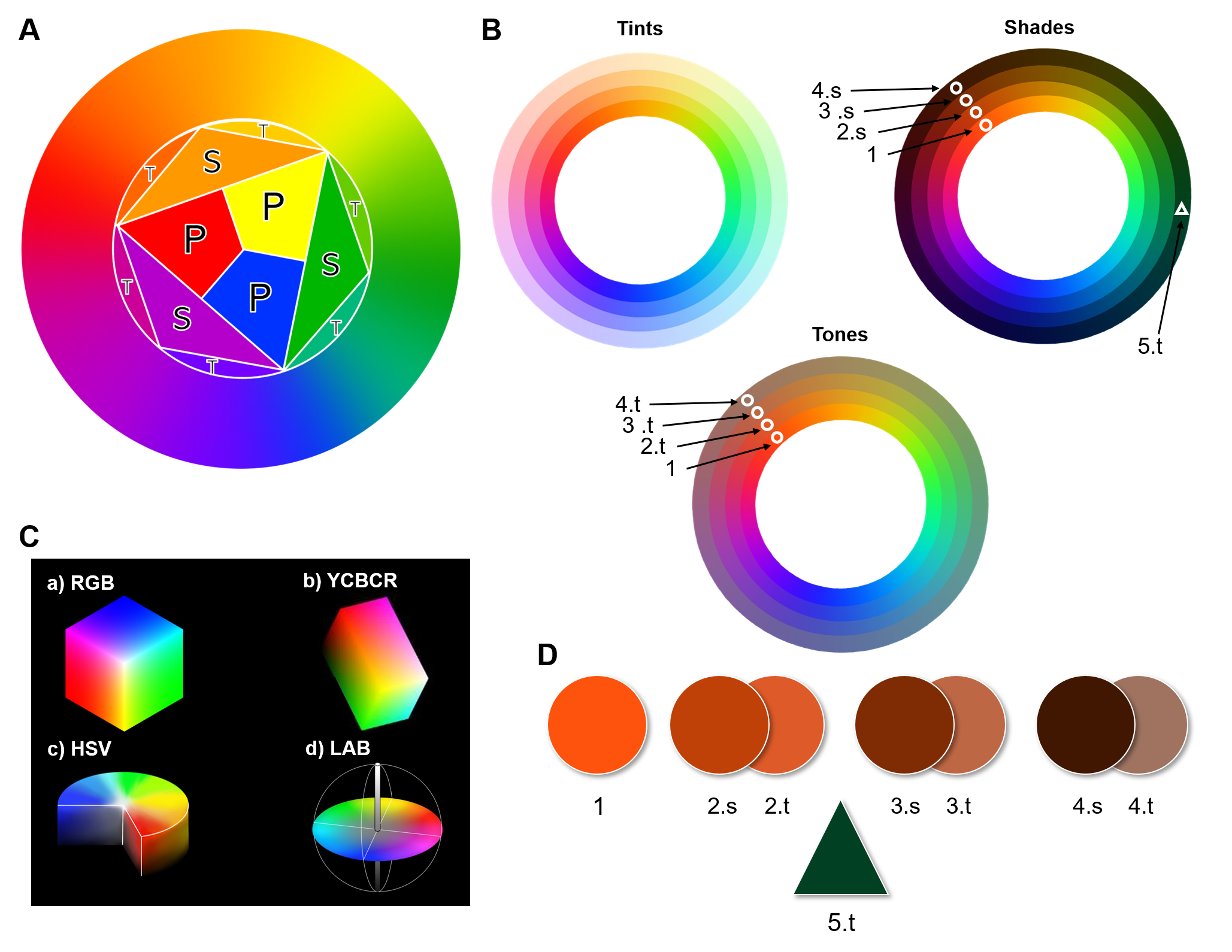}
  \caption{Color theory bases: (A) Representation of color wheel. The color wheel is obtained from the primary colors (P) Red, Yellow, and Blue. Blending the primary colors, we get the secondary (S) colors Green, Purple, and Orange. Finally, blending the secondary colors, we obtain the tertiary (T) colors: Yellow-Green, Blue-Green, Blue-Violet, and Red-Violet. Because colors are continuous, the best way to represent them is in a color wheel without boundaries, showing how they slowly evolve while mixing; (B) Depiction of the concept of tints, shades, and tones with the color wheel in (A). In every wheel, each ring going outward shows a 25\% increase in added white for tints, black for shades, and grey for tones. These modifications of the color wheel show the near-achromatic colors. The achromatic colors black, white, and grey are not shown.(C) The four main color spaces used in digital imaging: RGB, YCbCr, HSV, and CIELAB. (D) Color picker results displaying the colors within the shapes marked in the shades (s) and tones (t) modified color wheels. The circles show the effect of adding black (shades) or grey (tones) to orange. The triangle shows a dark shade of green called forest green. The obtained results show that what we understand as brown is not a pure color but rather shades of pure colors ranging from red to warm yellow.
}
  \label{fig:fig1}
\end{figure}

\subsection{Color theory and naming}
\label{subsec:theory}

Color theory is the science that explains how humans perceive color. Three color-making attributes are used to communicate the subjective experience of color: hue, hue purity (expressed as colorfulness, chroma, saturation, or dullness), and lightness or darkness.

To arrange the different colors for easier description, Newton devised the color or hue wheel in 1704 \cite{westfall1962development}. This wheel was created by superimposing the light spectrum ends (orange-red and violet)  to show the extra-spectral hues, which do not appear in the prismatic spectrum. This concept remained the same in the following centuries, as no significant modifications were needed. Figure \ref{fig:fig1}.A  shows how the hue wheel can also be understood as the progressive blending of the primary, secondary and tertiary colors. The first reference to this classification of colors dated from 1613 when the mathematician François d’Aguilon endorsed the medieval view of yellow, red, and blue as the basic or “noble” hues \cite{shamey2020d}. However, ancient Greek texts \cite{sorabji1972aristotle} show that early artists and dyers already utilized primary mixtures based on yellow, red, and blue colorants.

Nowadays, it is accepted that the primary colors are the parents of the color wheel: Red, Yellow, and Cyan. These colors are called primary because they cannot be created by mixing other colors. Blending two primary colors gives the secondary colors: Green, Orange, and Purple. Lastly, we obtain tertiary colors by mixing a primary color with its nearest secondary color. The six tertiary colors are Red-Orange, Yellow-Orange, Yellow-Green, Blue-Green, Blue-Violet, and Red-Violet. The primary, secondary and tertiary colors are called the Chromatic colors. This representation can be seen in the inner circle of Figure \ref{fig:fig1}.A.

It is important to note that some colors are not represented in the color wheel. This is because it only shows pure or fully chromatic colors. Modifications to these pure hues create the rest of the possible colors. The hues can be modified by changing the color, making attributes of purity and lightness with the achromatic or neutral colors: white, black, and gray. That is how we obtain tints, shades, and tones, respectively. Tints, shades, and tones are derived by adding white, black or gray to the pure hues, making them lighter, darker, or duller (less saturated), respectively, than the original colors. The whole variety of colors can be seen in the tints, shades, and tones images of Figure \ref{fig:fig1}.B. If we single out certain shades and tones (Figure \ref{fig:fig1}.D), we can better understand an important point of our work: the difference between pure hues and modifications. Triangle 5.t shows what is normally referred to as “forest green.” This deep shade of green is usually described simply as “dark green.”  Circles 1 to 4 show increasingly dark (1-4.s) shades and duller tones (1-4.t) of orange. When they are observed separately, it is clear that these colors are what we usually refer to as “brown.” Brown is technically a near-neutral dark-valued orange. However, in contrast to dark green, mostly no one would refer to brown as dark orange. This is a crucial point to understand because most research conducted on color detection and classification aims to find a unique space for brown as a pure hue instead of treating it as what it is: a modified hue, the same as dark green or dark blue \cite{mylonas2020coherence,mulholland2017identifying,mabberley2017painting}. This repeated inaccuracy in color research that contradicts even Goethe’s and Newton’s precise placement of brown as dark orange in their color wheels \cite{parkhurst1982invented,macevoy2005modern}, is likely to stem from the complicated gap between color theory, color perception, and color naming \cite{witzel2019misconceptions}.

The continuous nature of color makes classifying and naming individual shades problematic. Since the 17th century, several formal attempts at color naming and classification have been driven mainly by the boom of exploratory trips and expeditions. A. Boogert and R.Weller first ventured into this task, classifying colors according to object association with their \textit{Trait des Couleurs servant la Peinture l’eau }and the \textit{Tabula Colorum Physiologica}, respectively \cite{mclachlan2020visual}. M. Harris, based on Isaac Newton’s research on light, introduced his famous prismatic color wheel in the book \textit{The Natural System of Colors}. Following these ideas, in the early 18th century, T. Haenke and F. Bauer devised a 16 pages color palette to assist their botanical expeditions throughout South America \cite{mulholland2017identifying,mabberley2017painting}. In 1814, Abraham Gottlob Werne  and Patrick Syme published \textit{Warner’s Nomenclature of Colors}, which was used by artists and naturalists such as Charles Darwin in his 1832 expedition aboard the H.M.S. Beagle \cite{werner1821werner}. These works, however, relied on context and cultural cues. In 1912, Ridgway wrote that "the nomenclature of colors remains vague and, for practical purposes, meaningless, thereby seriously impeding progress in almost every branch of industry and research." His proposal, \textit{Color in a New Light}, did not prosper either, since his chosen pigments would change when exposed to light or humidity \cite{hamly1949robert}.
In the 1930s, the Inter-Society Color Council (ISCC) and the National Bureau of Standards  (NBS) first established the ISCC-NBS System of Color Designation to standardize color naming. The system was reformulated in 1955 according to the new Munsell coordinates. It defines three levels of color naming; the first level consists of 10 color terms, level 2 adds 16 intermediate categories, and level 3 uses brightness and pureness modifiers to obtain a final 267 color categories (see Figure 4 in evaluation section)  \cite{inter_societyCC}.
The closer we have gotten to standardization while considering different cultures and languages came in 1969 with Berlin and Kay. They researched color naming in different natural languages and found 11 basic color categories found in most of them: white, gray, black, yellow, orange, red, pink, purple, blue, green, and brown. In their work \textit{Basic Color Terms: Their Universality and Evolution}, they were able to link the universality of these terms to similar evolutionary processes \cite{berlin1991basic}. They also found that within these categories, the central regions would elicit higher consensus and fastest reaction times than the boundaries, where consensus tapers off, consistent with the continuous nature of color.

Berlin and Kay’s categorization and the ISCC–NBS System of Color Designation are the most used in digital imaging. Both approaches present problems that limit their practical use; Berlin and Kay’s color terms fail to combine successfully to create secondary and intersectional classes. Moreover, the experiments were conducted using exclusively high chroma levels, never testing less saturated colors. The ISCC–NBS names may seem arbitrary to different users that will not agree with the provided descriptions. Moreover, terms on levels 2 and 3 do not follow logical reasoning that could make them understandable to any human or machine, as demonstrated by Anthony E Moss in \cite{moss1989basic}.

Further research has proven that most people would use "simple" color terms instead, and only a small portion would use non-basic terms \cite{sturges1995locating}. These limitations are maintained throughout different cultures and languages \cite{witzel2019misconceptions,paggetti2011re,lillo2007locating,jraissati2018delving, abshire2016psychophysical,mylonas2022augmenting}.

\subsection{Color vision and color spaces }
\label{subsec:visandspac}
Color vision is the ability to perceive light with different wavelengths independently of light intensity. Perception of color is achieved thanks to our cone cells, a specialized kind of retinal cells in our eyes. According to their spectral sensitivities, these cells are subdivided into three subtypes: short, medium, and long cones. These three types correspond to red, blue, and green, respectively. Nevertheless, all humans have different cultural and contextual associations with color that vary our perceptions \cite{helmenstine2020visible,lee2008evolution,jacobs1996primate}.

The problems described in the previous paragraphs show that solving the semantic gap in the naming and classification of colors is a difficult task. Due to the difference in human color perception and contextualization, we would need experiments with a vast cohort of heterogeneous subjects to account for the differences in personal perception, eye morphology, cultural context, and professional bias \cite{mylonas2016augmenting,mylonas2014gender,lillo2007locating,abshire2016psychophysical,chamorro2016fuzzy}. However, we believe that a better approach is not to close the semantic gap but to build a sufficiently robust bridge capable of connecting both sides. Our goal is to find a color model that combines human perception and machine logic so that both understand the resulting names. Throughout the 20th century, several models were created to name color and represent it, matching our perception as closely as possible. These approximations are called color spaces.

The Munsell Color System, created in 1913 by Professor Albert H. Munsell to have a rational way to describe colors, is the most influential analog color space \cite{munsell1907color}. The United States Department of Agriculture (USDA) adopted this color space in the 1930s as the official color system for soil research. Its main idea is the distribution of colors along three perceptually uniform dimensions: hue, value, and chroma.  The placement of the colors was decided by measuring human virtual responses instead of trying to fit a predefined shape, thus obtaining an irregularly shaped cylindrical space that avoids distorting color relations \cite{munsell1912pigment}. In this space, hue is measured in degrees around horizontal circles, chroma goes radially outwards from the neutral vertical axis, and the value increases vertically from 0 (black) to 10 (white). The irregular shape of the Munsell color space allows for wider ranges of chroma for specific colors, which follows the physics of color stimuli and the nature of the human eye. This non-digital Color System is the most used in all research areas, particularly in psychology and soil research \cite{mylonas2016augmenting, pegalajar2020munsell}.

With the rise of computer operations, digital color spaces were created to suit the new tasks better. These color spaces can be classified into three categories: based on RGB primaries, cylindrical transformations, and human perception. A visual representation of these can be seen in Figure \ref{fig:fig1}.C.

sRGB -the main RGB primary color space- was created jointly by Hewlett-Packard and Microsoft in 1996 for Internet and display purposes, being the most common in digital applications. This color space is based on the RGB color model. Thus, it combines different proportions of Red, Blue, and Green narrow wavelengths to produce any color. Namely, colors are represented as a set of three coordinates corresponding to red, green, and blue values. Therefore, the sRGB color space can only represent shades that fall inside the triangular gamut created by the points corresponding to these three colors. Other issues include device dependency, discordance from human perception and interpretation, and a high correlation between the three primaries. These are common problems in all color spaces created using RGB primaries: sRGB, Adobe RGB and its variations, YCbCr, and YUV. These last two use linear combinations and gamma correction to separate the lightness and color or chroma signals in RGB.

As an alternative to the representations based on the RGB primaries, HSL and HSV cylindrical color spaces were designed by computer graphics researchers to be more aligned with their perception of color-making attributes. These spaces consist of a three-dimensional cylindrical transformation of the RGB primaries into a polar coordinate system, obtaining  Hue, Saturation, and Value (in HSV) or lightness (in HSL). The difference between HSV and HSL ultimately falls in how each represents brightness. Artists and computer vision researchers favor these spaces because their approximation to color seems more natural. However, these approaches present their own problems; HSV and HSL do not effectively separate tints into their three value components according to human perception.
Moreover, they ignore the complexity of color appearance and functioning by preserving RGB symmetries that confer the same lightness, chroma, and evenly spaced hues to the red, green, and blue primaries. Thus, these user-oriented color spaces trade perceptual relevance for computational speed \cite{poynton1997frequently}. This trade-off made sense when they first became popular for digital applications in the late nineties, but it is no longer a variable to consider.
The uniform color spaces based on human perception achieve the best compromise between complexity and accurate color representation. Of these, the most successful is CIELAB, created by the International Commission on Illumination. This color space implements the concepts introduced by the Munsell Color System. It ensures that the geometrical distance between any two points represents the perceived color differences without forcing the color relationships into strict geometrical shapes. This makes this color space more perceptually linear and decorrelated than the other alternatives. In CIE Lab, the luminance dimension we see in the RGB cube as a diagonal running from (0,0,0) to (1,1,1) is the vertical axis (L) of the space. The other axes are color channels A (green to red ratio) and B (yellow to blue ratio) \cite{seymour2020does}. Thanks to the above-presented properties, the A and B values can give a complete understanding of the chromatic content of a color independent of its luminance \cite{gershikov2007correlation, jin2013study}. The CIELAB color space provides a fully interpretable map of color behavior in each dimension and a precise depiction of shades, tints, and tones. However, it is not as widely used as HSL and HSV. While the lightness value is easy to understand, the coordinates corresponding to the green-to-red and yellow-to-blue ratios are more difficult to grasp for most people. The digital color spaces discussed in this section are displayed in Figure \ref{fig:fig1}.C.

\section{Proposal}
\label{sec:prop}
This paper introduces a new approach to pixel classification by color segmentation and naming. Our proposal entails important innovations concerning the state-of-the-art by applying color theory and fuzzy logic for color segmentation, identification, and classification.

Our method consists of two parts. First, the different colors presented in an image are identified and classified into 12 color classes. Second, each identified color is named based on color theory and fuzzy-inspired multi-label classification to provide a bias-free description.

For the first part, we exploit the rationale behind classical color theory to create a reduced, two-dimensional color space based on CIELAB. This allows us to provide accurate color identification and segmentation while avoiding perception bias and eliminating the need for a wide cohort of labels. Color theory principles ensure that results are coherent, without supervision, as opposed to previous works that mapped color terms to pure hues, misclassified colors by labeling them with opposing areas of the color wheel, or failing to account for shades and tonality. This last item is especially important in the case of browns, a group of colors that our method can successfully identify.

For the second part, we introduce a mathematical approach for describing the different shades, tones, and tints present in an image. This is done using fuzzy logic and multi-label systems to avoid bias introduced by human labels and prototypes, thus eliminating contextual and cultural errors.

Finally, we provide a complete and easy-to-understand summary of the input image, detailing its color content in full, the composition of each shade, tone, and tint, and all the different color segmentation masks. This information can then be applied in further analysis in soft labeling, classification, and training tasks.

\section{State of the art}
\label{sec:sota}

Much research has been devoted to color verbalization and understanding. These studies aim to find a suitable mapping between the digital representation of colors and the language employed by humans when referring to color in an attempt to close the semantic gap successfully \cite{abshire2016psychophysical,mylonas2022augmenting,yazici2018color}. This task is arduous because the human understanding of color is entirely subjective. It depends on age, context clues, professional bias, eye morphology, etc. Therefore, the only way of successfully closing this gap is by running experiments to obtain an appropriate representation of color categories. However, this would require researchers to collect a sufficiently heterogeneous cohort that would consider every possible experience of color, which is impractical. 

In a realistic setting, the primary methods used for color segmentation and classification fall under two categories: quantization of the space and fuzzy models.

 Color space quantization is based on clustering. Typically, clusters are computed either by applying color thresholding or nearest neighbor classification. More recently, deep learning-based clustering approaches have obtained improved results. Color thresholding divides the space into segments defined by mostly-linear boundaries that need to be decided. This step has been solved using machine learning techniques, which require a priori knowledge of the labels of the color categories and training process, or by complex deep learning optimization algorithms \cite{xing2020improved,bao2019novel,liang2019modified}. Alternatively, classification based on nearest neighbors finds the \textit{k} closest neighbors to any given pixel in a set of predefined color classes. Like the previous method, this approach is severely limited by the necessary definition of the clusters and boundaries, which depend entirely on the training set \cite{TIAN2019104962,9288648}. This fact makes both solutions highly problem-dependent and difficult to adapt to other tasks, although some recent research provides more robust results \cite{basar2020unsupervised}. Overall, color space quantization methods mostly focus on object segmentation and fail to classify the detected shades.

 Fuzzy models have been introduced recently to tackle the notoriously ill-defined boundaries between colors. Fuzzy set theory is closer to the human cognitive process of color categorization than crisp color partitions. Because of this, several research works have applied fuzzy logic to successfully obtain a correspondence between the computational representation of colors and human perception. Within this approach, two subgroups of methods stand out: 1) those that rely on geometrically regular membership functions obtained from observer experiments;  \cite{dadwal2012estimate, abshire2016psychophysical,shamir2006human,leon2020big}, 2) those going beyond and formalize fuzzy color spaces from human-obtained prototypes \cite{mylonas2022augmenting,chamorro2016fuzzy,chamorro2019granular,chamorro2021path, mengibar2022learning}. Methods in the first subgroup are problematic because they rely on color spaces that have arbitrary distances between color categories (HSV, HSL, RGB triplets) and require many observers. Moreover, these proposals focus on a limited number of color terms obtained from subsets of a single cultural environment (like Berlin and Kay). Methods in the second subgroup achieve the best possible solution to close the semantic gap in specific situations. However, their dependence on prototypes and seeds limits their robustness in color segmentation. Collectively, methods based on fuzzy logic for classifying color pixels focus more on naming the colors than on color segmentation.

Nevertheless, the problem all previous methods have in common is the disregard for color theory fundamentals. Most of the approaches mentioned above are based on misinformed notions of what colors are. These inaccuracies are often due to the confusion between pure hues and color terms. Basic Color Terms (BCTs as introduced by Sturges \& Whitfield in 1991) include 11 categories: Black, White, Red, Green, Yellow, Blue, Brown, Orange, Pink, Purple, and Grey \cite{berlin1991basic,sturges1995locating}. Since then, many research works have aimed to locate these color terms in the color spaces as if they were pure hues. This becomes particularly problematic in the case of brown \cite{abshire2016psychophysical} for the reasons stated previously.

\section{Methods}
\label{sec:methods}

To create the proposed method, we approach the problem in two steps. First, we solved automatic color identification and segmentation. Namely, we studied the color wheel and obtained the pure hue bases in the CIELAB color space. We named this step “Geometric knowledge generation.” Second, we used fuzzy logic notions and art theory to name and describe the identified colors. This step is called “Fuzzy space definition and multi-label classification for color naming and description.”

\subsection{Geometric knowledge generation}
\label{subsec:geometric_knowledge_generation}

\begin{figure}[ht!]
  \centering
  \includegraphics[scale=0.5]{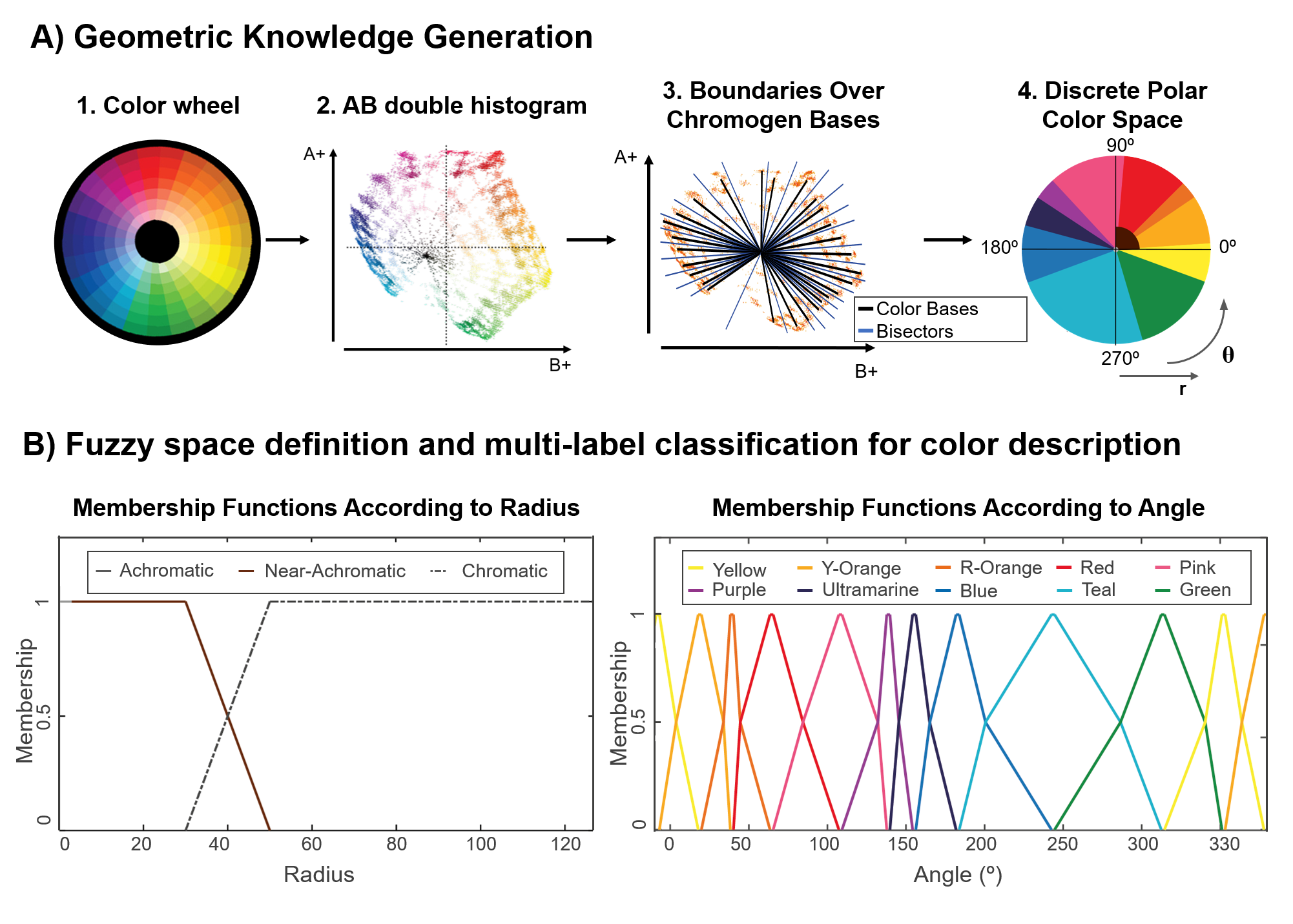}
  \caption{Two main steps of ABANICCO: (A) Geometric knowledge generation. We identify in the color wheel the localization of the different pure hues, shades, and tints within the reduced AB space of CIELAB. Using bisectors, the best boundaries between pure hues are found. With this, we obtain a discrete polar color space divided into 12 different color categories: Pink, Red, Red-Orange, Yellow-Orange, Yellow, Green, Teal, Blue, Ultramarine, Purple, Brown, and Achromatic. The first 10 categories describe hue depending on the angle. The last two describe chromaticity and depend on the radius. 
(B)  Fuzzy space definition and multi-label classification for color description. We formalize fuzzy rules to translate the obtained discrete color polar space back into a continuous space. Based on the bisectors employed for the discretization of the space, we can define areas of absolute chromogen certainty and gradients of membership to the adjacent colors on both sides. This is done in both the radius (accounting for chromaticity) and the angle (accounting for hue).}
  \label{fig:fig2}
\end{figure}

Our final goal is to create a simplified color space for color classification and naming. As explained in Section \ref{subsec:visandspac}, we chose to work with CIELAB color space due to its advantageous properties.

The first step was to find the exact location of the pure hues present in the color wheel within the CIELAB color space. Since color space dimensions are decorrelated, the luminance component is not needed to understand a given color's chromatic content. This meant that we could discard the luminance dimension and work on the reduced AB two-dimensional space. To find the location of the main crisp or pure hues in this reduced color space, we selected an example of a complete color wheel. We calculated the double AB histogram, showing the location of the different crisp hues. The resulting geometrical distribution apparently radiates from a center point, as shown in Figure \ref{fig:fig2}.A. The skeletonization of the double histogram provides endpoints and branches following a circular path. Due to their disposition, the lines between the center of the dual histogram and each endpoint can be used to represent the bases of the pure chromogens.

\begin{figure}[]
  \centering
  \includegraphics[scale=0.5]{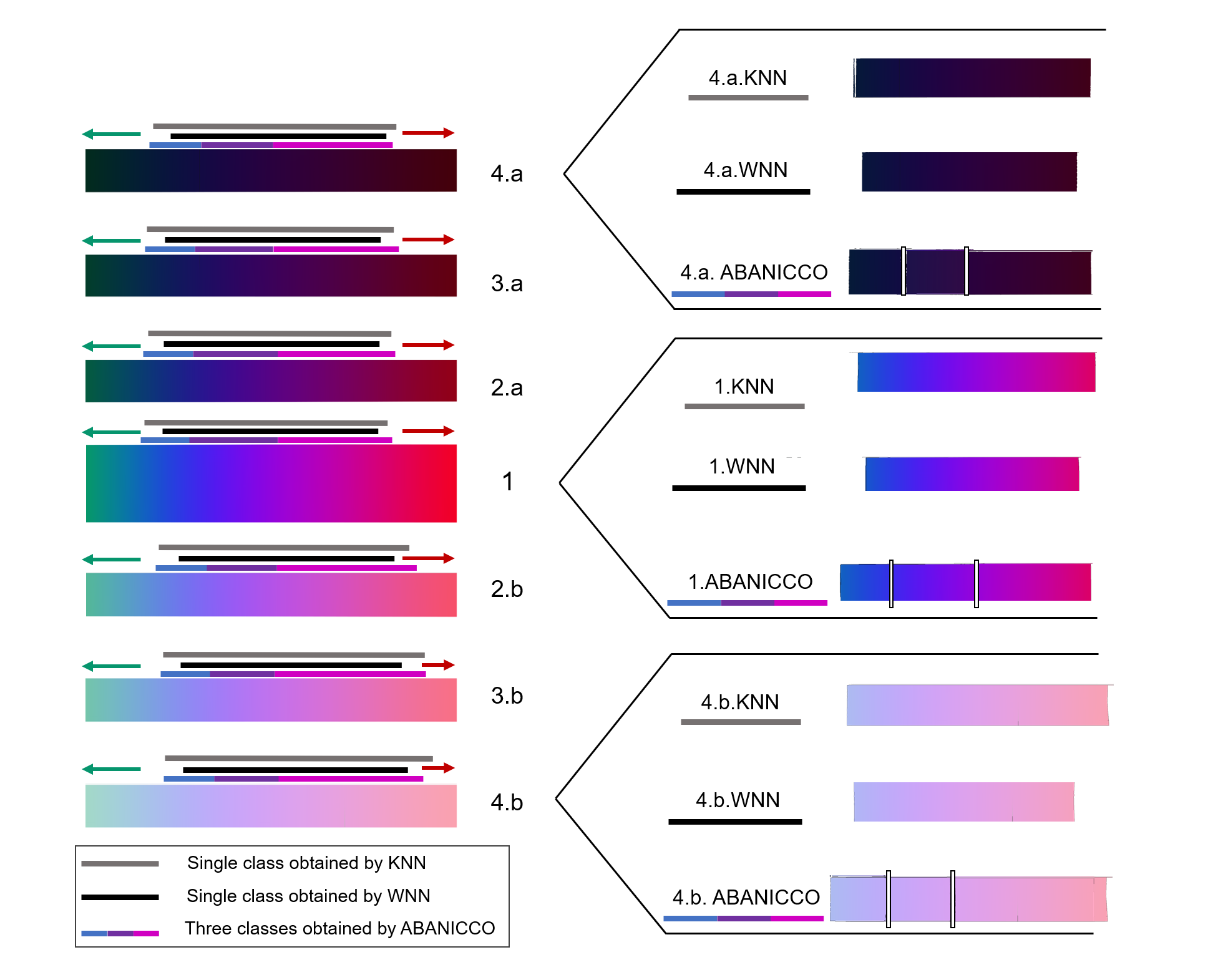}
  \caption{Comparison of color classification by machine learning methods and our method: This evaluation was carried out in 7 rectangles depicting the concept of shadows and tones of a portion of the color wheel. Rectangle 1 shows the portion of the color wheel from bluish-green to red. Rectangles above this have been obtained by adding an increasing quantity of black to represent shades (Rectangles 2-4 a). Rectangles below have been obtained by adding an increasing quantity of grey to represent tones (Rectangles 2-4 b). The figure displays the results of color classification over the region covering the purplish-blue and pinkish-red colors. However, the region's exact limits and length are decided by each algorithm.
Above each rectangle, three different lines measure the length of the classes obtained by each method for this portion of the color wheel. The solid gray line is the classification carried out by the k-nearest neighbors algorithm (KNN); the solid black line by the Wide Neural Network (WNN); the tri-color line by our method. A solid color line represents both Machine Learning methods because they obtain just one class from colors ranging from blue to pinkish-red, unable to differentiate between blues, purples, and pinks. In contrast, our method is represented using three colors because it obtains three different classes in all the experiments regardless of tonality. To both sides of these lines are arrows indicating the classes obtained before and after this section. To the left, a green arrow indicates that the previous class would be Green. To the right, a red arrow indicates that the following class would be Red. However, after introducing the option of near-achromatics, ABANICCO would determine that portion as Brown in the case of rectangles 2-4.a. 
The actual segmentation performed by the three methods on rectangles 4.a, 1, and 4.b is shown to the right of them. The separation between classes of our method has been marked with white lines for clarity.}
  \label{fig:fig3}
\end{figure}

We considered two options to locate the boundaries between the crisp color categories: angle bisectors and classification methods. Both were tried out before moving to the next step to decide on the best approach. The classification methods were more complex to use, needed training, took longer to provide results, and the obtained classifications were unsatisfactory both in resulting boundaries and classes created. An example of both approaches segmenting a gradient of colors composed of blues, purples, and pinks is shown in Figure \ref{fig:fig3}. We observe how the machine learning classifiers fail to create appropriate boundaries between the three present categories, each generating a single class of arbitrary length instead of the three expected classes of set length obtained with the bisectors approach. This clarifies why we chose the bisectors approach. The experiments carried out to determine the best method are described in Section \ref{sec:Supplementary}. 

Computing angle bisectors, we obtained precise color regions that were then grouped according to color perception, CIELAB construction, and Berlin and Kay theories \cite{berlin1991basic, gershikov2007correlation} in 10 groups: Green, Yellow, Light Orange, Deep Orange, Red, Pink, Purple, Ultramarine, Blue, and Teal.

It is important to note that neutrals (black, white, and grays) and near-neutral brown are not present in this classification as they are not pure chromatic colors.  In the CIELab color space, the pure hues are located far from the center and radially distributed. All modifications of these hues then fall gradually closer to the center, with the shades having lower luminance than the tints and the achromatic colors falling in a gradient aligned on the L axis. This can be seen in the CIELAB diagram of Figure \ref{fig:fig1}.C. In our reduced AB space, this means that the neutral colors black, white, and gray are located in the center of the space. The shades and tones of orange, commonly named brown, are located close to the center within the red, orange, and yellow areas. Figure \ref{fig:fig2} shows how this area corresponds to the lower-left corner of the upper right quadrant in the AB space (consistent with the mixed range of yellows and reds), close to the L axis (due to their near-achromacity). Therefore, brown shades can be aligned with any pure chromogen bases between red and warm yellow depending on the angle and distance from the center.

So, finally, we include two additional areas in our simplified color space that depend on the distance from the center: The first one is the circular area of radius r1 (Critical Achromatic Radius) corresponding to the neutral pixels centered on the L axis: the whites, grays, and blacks in the image. The second area is a quadrant of radius r2 (Critical Near-Achromatic Radius) corresponding to the positive values of A and B in the 2D colorspace and comprising all shades of brown. The radii r1 and r2 correspond to the first and second sets of branching points in the skeletonized double histogram (supplementary material).

In this manner, a new, simplified color space is created, capable of classifying colors into 12 categories: Pink, Red, Red-Orange, Yellow-Orange, Yellow, Green, Teal, Blue, Ultramarine, Purple, Brown, and Achromatic, as seen in Figure \ref{fig:fig2}. Within our pipeline, this new color space is used in the pixel classification section: the RGB input image is converted into CIELAB, and the AB values are translated into polar coordinates (radius and angle). Each pixel is classified into one of the 12 categories with these coordinates. The categories are consistent with the unbroken areas obtained using angle bisectors over the color bases. This classification depends solely on the angle for the first ten categories, the neutral category's radius, and the Brown category's angle and radius.

To classify colors with this method, the pixel values of the input image are converted from RGB to CIELAB, and the resulting AB coordinates are converted to polar $(r, \theta)$, identifying to what color that pixel belongs.

\subsection{Fuzzy color space formalization and color naming}
\label{subsec:fuzzy_color_formalization}

As described in the previous section, using the angle bisectors on the double AB histogram, we can successfully detect and separate different chromogens, accurately segmenting the detected colors. In this Section, to approach the goal of defining an unbiased continuous space for color naming and classification, we used the principles of fuzzy logic applied to multi-labeling systems. In this way, we use trapezoidal heterogeneous membership functions to represent fuzzy colors. Namely, we define  as areas of absolute chromogen certainty the points lying on top of the bases and define gradients to the colors of both sides. The method was inspired by \cite{chamorro2016fuzzy,chamorro2019granular,chamorro2021path,mengibar2022learning,chamorro2021referring,forcen2017adding}. 

We defined the set of membership functions using polar coordinates, as shown in Figure \ref{fig:fig1}.B, and trapezoidal piecewise polynomials to maintain continuity, simplifying the problem and matching the geometric angular analysis. For the angular membership, linear piecewise functions were fitted for each color class so that they would output 1 when the angle of the pixel was in the middle (or peak) of the color class, 0.5 in the boundaries, and 0 when over the peak of the next or previous color class as given by
\begin{eqfloat}
\begin{equation}
\label{eq1}
	f(\theta,a,b,c,d,e,g)=%
    \begin{cases}
        0 &\text{if $\theta$ $\leq$ a} \\
        \frac{\theta -a}{2(b-a)} &\text{if a $<$ $\theta$ $<$ b}\\
        \frac{\theta -2b +c}{2(c-b)} &\text{if b $<$ $\theta$ $<$ c}\\
        1&\text{if c $<$ $\theta$ $<$ d}\\
        \frac{\theta -2e +d}{2(d-e)} &\text{if d $<$ $\theta$ $<$ e}\\
        \frac{\theta -g}{2(e-g)} &\text{if e $<$ $\theta$ $<$ g}\\
        0 &\text{if  $\theta$ $\geq$ g},\\
    \end{cases}
\end{equation}
\end{eqfloat}
where $a$ is the peak for the previous color class, $b$ and $e$ are the angles corresponding to the boundaries, $c$ and $d$ are the angles adjacent to the peak of that color class, and $g$ corresponds to the peak of the next color class.

This was done using the geometric knowledge learned from the AB reduced space. The radius membership functions were created to differentiate between achromatic (scale of grays), near-achromatic (brown), and pure colors. The function defining the achromatic area was modeled with two linear splines, as given in
\begin{eqfloat}
\begin{equation}
    \label{eq2}
    f_{Achromatic}(r,r_{1})=%
        \begin{cases}
            1 &\text{if r $\leq$ r\textsubscript{1}} \\
            0 &\text{if r $>$ r\textsubscript{1}}.
    
    \end{cases}
\end{equation}
\end{eqfloat}

While three linear splines were used to define the near-achromatic area
\begin{eqfloat}
\begin{equation}
\label{eq3}
    f_{Near-Achromatic}(r,r_{1},r_{2}',r_{3})=%
    \begin{cases}
        0 &\text{if r $\leq$ r\textsubscript{1}} \\
        1 &\text{if r\textsubscript{1} $<$ r $<$ r\textsubscript{2}'}\\
        \frac{r-r\textsubscript{3}}{r\textsubscript{2}'-r\textsubscript{3}} &\text{if r\textsubscript{2}' $<$ r $<$ r\textsubscript{3}}\\
        0&\text{if r$\geq$r\textsubscript{3}},
    
    \end{cases}
\end{equation}
\end{eqfloat}

and the chromatic area

\begin{eqfloat}
\begin{equation}
\label{eq4}
    f_{Chromatic}(r,r_{2}',r_{3})=%
    \begin{cases}
        0 &\text{if r $\leq$ r\textsubscript{2}'} \\
        \frac{r-r\textsubscript{2}'}{r\textsubscript{2}'-r\textsubscript{3}} &\text{if r\textsubscript{2}' $<$ r $<$ r\textsubscript{3}}\\
        1&\text{if r $\geq$ r\textsubscript{3}}. 
    \end{cases}
\end{equation}
\end{eqfloat}

In the above equations, $r\textsubscript{1}$ is the Critical Achromatic Radius, $r\textsubscript{2}'$ is the Critical Near-Achromatic Radius, and $r\textsubscript{3}$ is the Critical Chromatic Radius, obtained from the skeletonized histogram.

It can be appreciated in Figure \ref{fig:fig2} that while the splines for the near-achromatic and achromatic areas show a similar logic to the angular membership functions, with a clear gradient between the two classes, that is not the case for the achromatic area. To accommodate for this, the Critical Achromatic Radius (r1) was used as defined in the previous section. Namely, all colors with radii lower than r1 are classified as achromatic. However, the Critical Near-Achromatic Radius (r2) was used as the point at which the membership function would output 0.5 for near-achromatic and chromatic. Thus, two new points were created: the New Critical Near-Achromatic Radius (r2’) and the Critical Chromatic Radius (r3), both separated 5 points from the original Critical Near-Achromatic Radius (r2).

With this approach, we get an exhaustive analysis of each shade present in the input image. We obtain a list of the ten main shades from each color category. Each shade is then described as a percentual combination of the main color categories that are required to create it. Thus, a shade with polar coordinates R=41, Theta=8º will be named "44\% Brown, 36\% Light Orange, 20\% Yellow," corresponding to a yellowish light beige. We believe this naming technique based on mathematical principles provides an excellent solution to the semantic gap, representing an ideal halfway point between human understanding and computer representation of colors.

\section{Results and Discussion}
\label{sec:results_discussion}

We have introduced ABANICCO, a novel color classification, naming, and segmentation method. In this section, we thoroughly evaluate its performance with other state-of-the-art methodologies and standards.
As previously mentioned, we want to emphasize that a perfect ground truth against which we may compare our model does not exist because of human perception, context, and color interpretation variability. Nevertheless, we manage to devise a quantitative evaluation that goes further from a visual assessment of the results. In particular, we first evaluate the method classification and naming accuracy and then its segmentation accuracy. 

\subsection{Classification and Naming}
\label{subsec:classification_naming}

\begin{figure}[]
  \centering
  \includegraphics[scale=0.45]{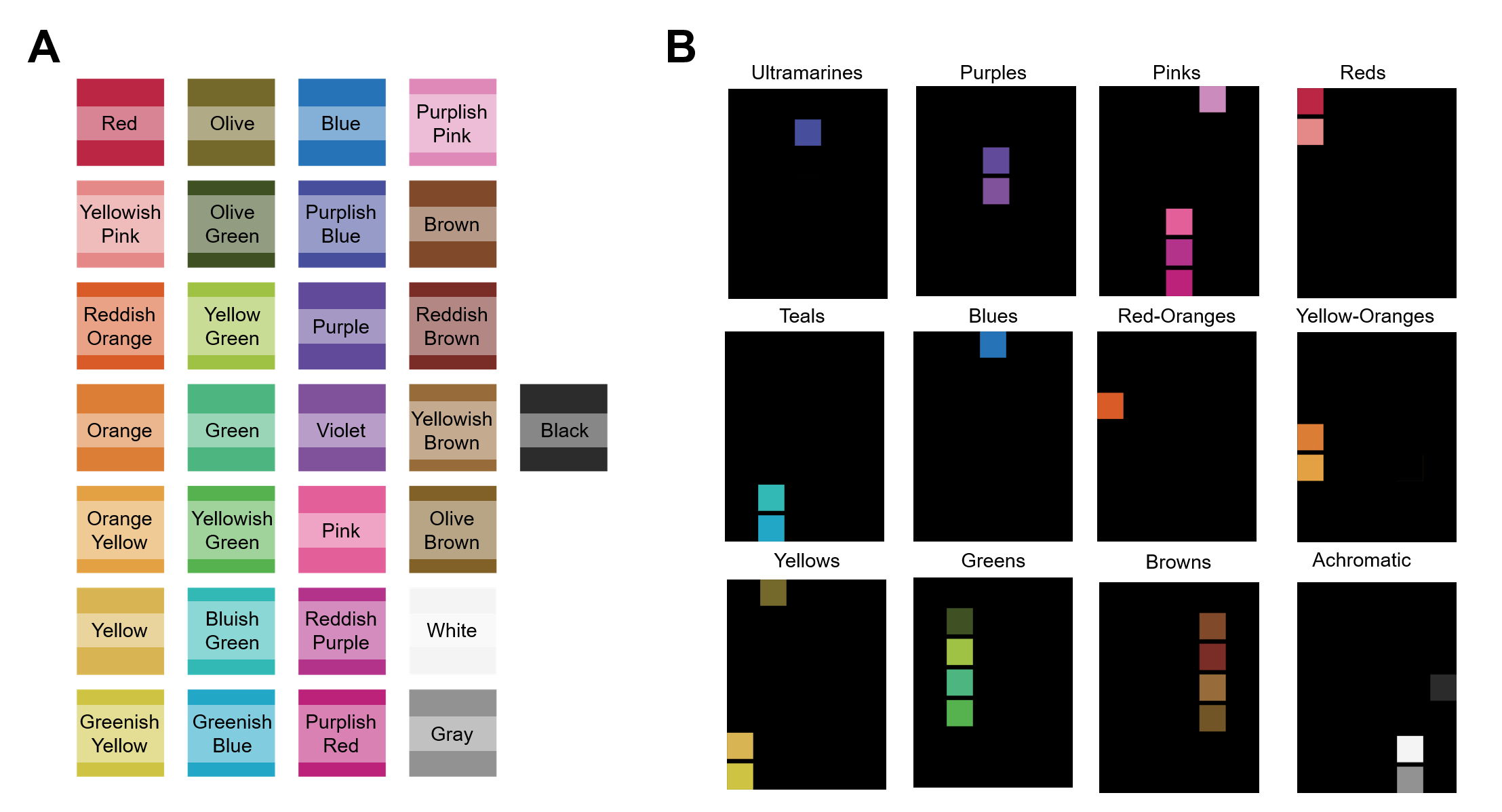}
  \caption{Evaluation of our method with ISCC-NBS system of color categories: (A) Level 2 of ISCC-NBS system with each color’s name superimposed over its corresponding square. (B) ABANICCO Classification of image A into the resulting 12 categories.}
  \label{fig:fig4}
\end{figure}

To evaluate ABANICCO for classification and naming accuracy, we apply our model to the official ISCC NBS color classification level 2 \cite{inter_societyCC}. This includes a set of 29 colors made up of 10 hues, 3 neutrals, and 16 compound color names. The model's classification and naming performance is summarized in Table \ref{tab:table} and Figure \ref{fig:fig4} . Specifically, Figure \ref{fig:fig4} reflects the results regarding color classification while Table \ref{tab:table} reports the naming comparison performance.
This experiment proves that our method can successfully classify the given shades into the expected categories and provide complete, coherent labeling per the ISCC–NBS system names. However, some of the names in the ISCC-NBS system are senseless from a color-wheel perspective. Examples of this are ’Yellowish pink’ and ’Reddish Purple,’ which combine two non-adjoining categories skipping the middle ’Red’ and ’Pink,’ respectively.  Our method solves these nonlinearities by classifying those colors within the skipped middle categories. These two scenarios (marked in bold in Table \ref{tab:table}) are the only ones reporting discrepancies between our method and the ISCC-NBS system.

\begin{table}[]
 \caption{Full Analysis of each shade of level 2 in ISCC–NBS extended set.}

  \centering
  \begin{tabular}{l|l|p{65mm}}
    \toprule
    \multicolumn{3}{c}{}                   \\

    \textbf{ISCC–NBS System Name} & \textbf{Algorithm's Classification} & \textbf{Algorithm's Full Description} \\
    \midrule
    Red (R) &Red &86.09 \%   Red and 13.91 \%   Pink      \\
Yellowish Pink (Ypk) &\textbf{Red} &58.77 \%   Brown, 40.04 \%   Red, and 1.19 \%   Pink     \\
Reddish Orange (Ro) &Deep Orange &78.67 \%   Deep Orange and 21.33 \%   Red     \\
Orange (O) &Light Orange &62.69 \%   Light Orange and 37.31 \%   Deep Orange     \\
Orange Yellow (OY) &Light Orange &93.41 \%   Light Orange and 6.59 \%   Yellow     \\
Yellow (Y) &Yellow &55.69 \%   Yellow and 44.31 \%   Light Orange     \\
Greenish Yellow (Gy) &Yellow &97.24 \%   Yellow and 2.76 \%   Green     \\
Olive (Ol) &Yellow &97.61 \%   Yellow and 2.39 \%   Light Orange     \\
Olive Green (Olgr) &Green &74.38 \%   Green and 25.62 \%   Yellow     \\
Yellow-green (YG) &Green &64.75 \%   Green and 35.245 \%   Yellow     \\
Green (G) &Green &62.73 \%   Green and 37.27 \%   Teal     \\
Yellowish Green (Yg) &Green &98.27 \%   Green and 1.73 \%   Teal     \\
Bluish Green (Bg) &Teal &86.26 \%   Teal and 13.71 \%   Green     \\
Greenish Blue (Gb) &Teal &75.88 \%   Teal and 24.12 \%   Blue     \\
Blue (B) &Blue &88.12 \%   Blue and 11.78 \%   Ultramarine     \\
Purplish Blue (Pb) &Ultramarine &100 \%   Ultramarine     \\
Violet (V) &Purple &57.27 \%   Purple and 42.73 \%   Ultramarine     \\
Purple (P) &Purple &73.47 \%   Purple and 26.53 \%   Pink     \\
Pink (P) &Pink &74.31 \%   Pink and 25.69 \%   Red     \\
Reddish Purple (Rp) &\textbf{Pink} &100 \%   Pink     \\
Purplish Red (Pr) &Pink &82.42 \%   Pink and 17.58 \%   Red     \\
Purplish Pink (Ppk) &Pink &85.01 \%   Pink and 14.99 \%   Purple     \\
Brown (Br) &Brown &77.72 \%   Brown, 7.16 \%   Light Orange, and 15.12 \%   Deep Orange     \\
Reddish Brown (Rbr) &Brown &50.10 \%   Brown, 8.28 \%   Deep Orange, and 41.62 \%   Red     \\
Yellowish Brown (Ybr) &Brown &69 \%   Brown and 31 \%   Light Orange     \\
Olive Brown (Olbr) &Brown &64.40 \%   Brown, 26 \%   Light Orange, and 9.60 \%   Yellow     \\
White (Wh) &Neutral &67.50 \%   Yellow and 32.50 \%   Light Orange     \\
Gray (Gy) &Neutral &67.50 \%   Yellow and 32.50 \%   Light Orange     \\
Black (Bk) &Neutral &67.50 \%   Yellow and 32.50 \%   Light Orange     \\

    \bottomrule
  \end{tabular}
  \label{tab:table}
\end{table}

\subsection{Segmentation}
\label{subsec:segmentation}

\begin{figure}[]
  \centering
  \includegraphics[scale=0.12]{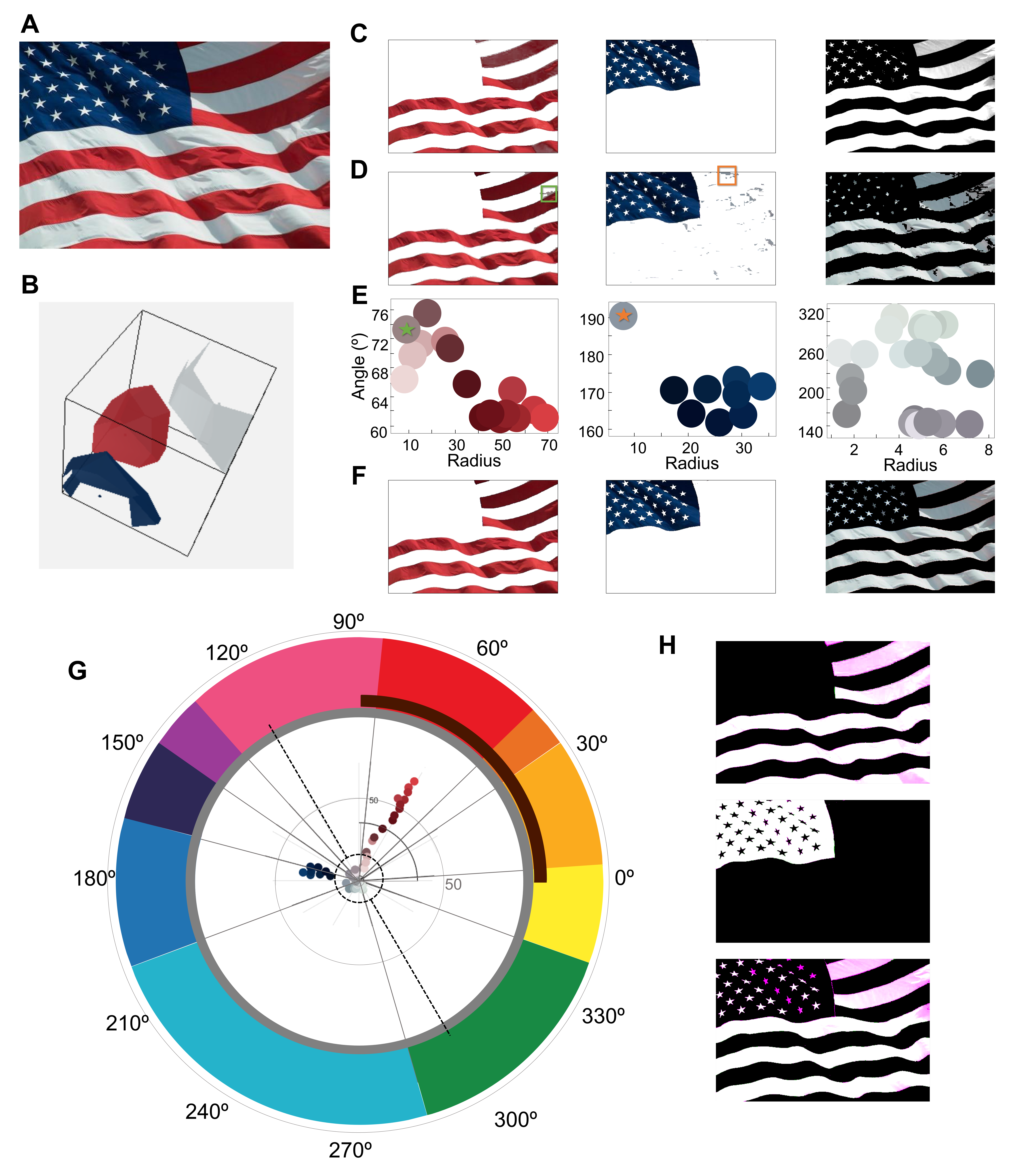}
  \caption{Comparison of segmentation with our method (ABANICCO) versus fuzzy method in \cite{mengibar2022learning}:
(A) Original natural image with a flag waving.
(B) Fuzzy color space built from extracted prototypes using the fuzzy method.
(C) Row of results (red, blue, and white mapping) of the segmentation using supervised prototypes with the proposal in \cite{mengibar2022learning}. In this and the successive rows, the “White” mapping background is shown in black to visualize results better. This technique uses membership degrees to represent the segmentation; thus, lower saturation corresponds to lower class membership certainty.
(D) Row of results (red, blue, and white category) of the segmentation using ABANICCO. The green and orange squares show areas where our method fails to segment the stripes accurately. 
(E) Colors within the red, blue, and light categories found by ABANICCO. The stars mark the color of the areas of failed segmentation in (D): The green (orange) star marks the color of the area within the green (orange) rectangle. 
(F) Row of segmentation results (red, blue, and white category) using ABANICCO after adjusting the boundaries. 
(G) Polar description of the image with the original boundaries in gray and the revised boundaries shown with dotted black lines. 
(H) Overlap of the masks obtained in row F and those obtained with \cite{mengibar2022learning}. The different shades of pink indicate where the method in \cite{mengibar2022learning} outputs an uncertain membership to that class. The stronger the pink, the lower the membership. In these images we can see how their method starts to fail in areas where the differences in illumination change the expected colors: dark red (brown), dark white (gray).}
  \label{fig:fig5}
\end{figure}

\begin{figure}[]
  \centering
  \includegraphics[scale=0.35]{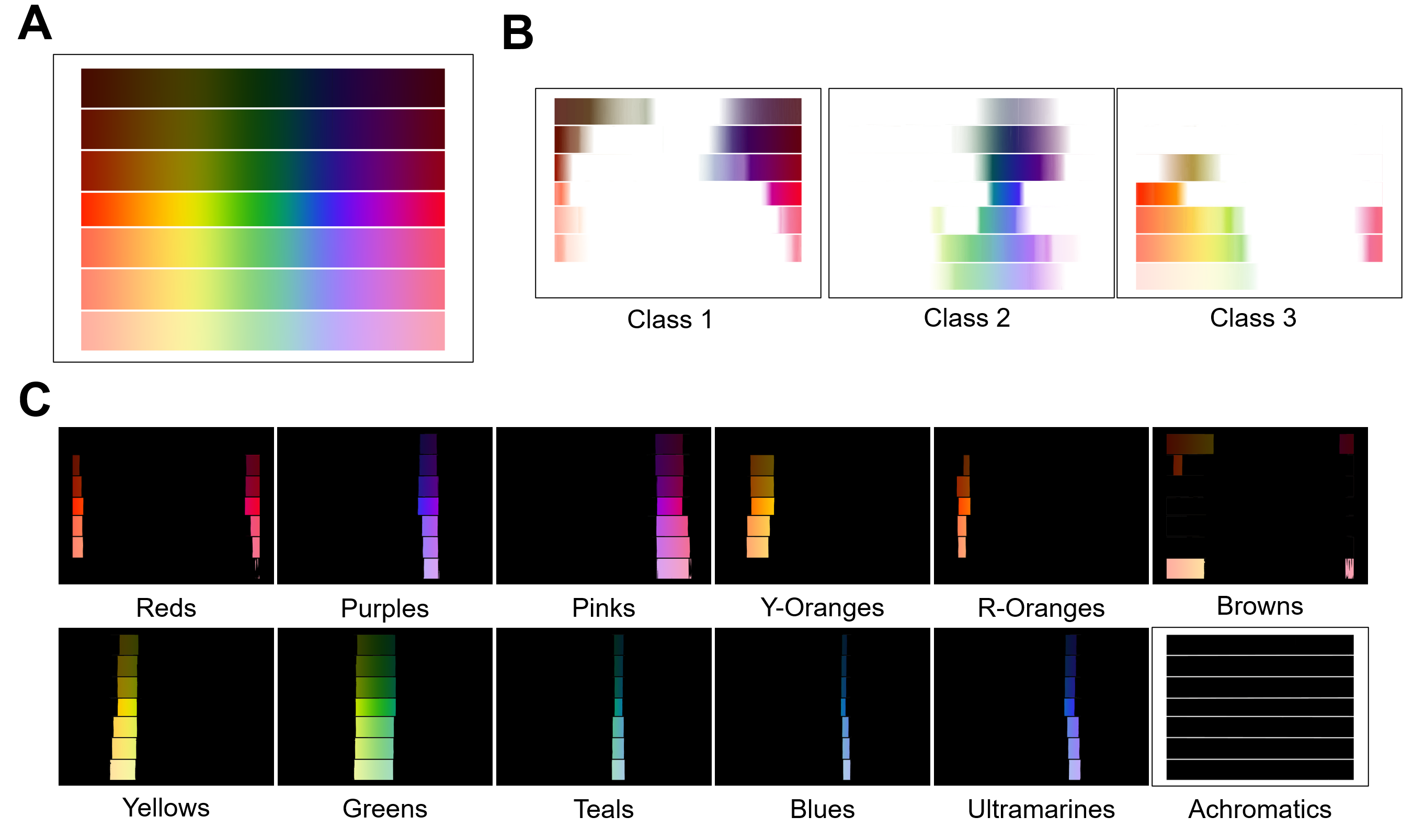}
  \caption{Comparison of segmentation with our method versus fuzzy method in \cite{mengibar2022learning} using a simplified version of the image shown in  Figure \ref{fig:fig1}:
(A) The middle rectangle shows the unraveled color wheel. The rectangles above (below) add an increasing quantity of black (grey) to represent shades (tones). 
(B) Results of classification using Fuzzy color spaces with 3 arbitrary classes. The lack of labels and the gradual transitions between colors result in membership maps with small areas of absolute membership and unclear class separations.
(C) Results of color classification using ABANICCO. Our method automatically divides the image into 12 classes with clear boundaries, corresponding to the main hues present plus the near-achromatic brown shades and tones and the achromatic white and black (empty for this particular image).}
  \label{fig:fig6}
\end{figure}

Once ABANICCO classification and naming accuracy have been validated,  we proceed to evaluate our model’s segmentation accuracy. As previously mentioned, ABANICCO segments color without any human input. This requirement was particularly important to avoid human bias in color identification. We choose to compare the model we proposed with the most representative method of the state-of-the-art for color classification, published very recently. In their work, M. Mengíbar-Rodrigez et al. \cite{mengibar2022learning} train a Self-Organizing Map (SOM) to obtain a suitable set of prototypes upon which to build a fuzzy color space for each image.
In contrast with ABANICCO, their method requires the user to decide on the number of categories. The model returns as output a map of the membership degree of each pixel to the studied category, which they represent by modifying the degree of transparency of the segmented images. We use their results to evaluate our model performance. Specifically, they segment a flag into three classes compared to a human-generated ground truth of 12,000 color samples. Figure \ref{fig:fig5}  shows the comparison between the results of both models. The image selected presents clear boundaries between colors far away from each other in the color wheel, which can be easily appreciated in the polar description of the image. Additionally, we compare the two methods using an image with progressive transitions between colors adjacent to each other in the color wheel in Figure \ref{fig:fig6}.

It is important to remind that ABANICCO has been created for color segmentation and classification, which does not necessarily imply accurate object segmentation. However, the polar description of the boundaries can be modified after the initial segmentation to better match this particular task. We illustrate this in Figure \ref{fig:fig5}, where the input image is a waving flag with white, blue, and red areas presenting diverse lightning due to movement. When ABANICCO is applied, the initial result does not give a perfect segmentation of the objects: stripes and stars (row D). On closer examination, it becomes clear that the segmentation errors are actually a result of ABANICCO focus on color classification. Namely, the areas in which either blue or red are reflected onto the white stripes are classified as blue and red, respectively. The actual color of these areas can be seen marked with a star in the full colors summary of row E. Therefore, the segmentation using ABANICCO is accurate in terms of color but not in terms of object edges. However, using the information given by the shade classification (Row E) and polar description (G), the user can easily modify the class boundaries and classes to best match the application (row F). In fact, when we compare our modified segmentation with the one obtained using \cite{mengibar2022learning}, we can see that the baseline method has problems classifying the different shades of blue and also the red color when it appears darker due to lighting and movement (Figure \ref{fig:fig5} H). This is most patent in the upper right corner of the image. Thus, our method can easily go from pure color classification to object segmentation by allowing the user to make informed decisions concerning class boundaries.

This particular image of a flag is easy to segment for clustering-based methods such as \cite{mengibar2022learning} because there are few colors present, and they are far from each other in the color wheel. To contrast this, we conducted the same evaluation on another image with no clear edges, the whole spectrum of hues of the color wheel, and different shades and tones of said hues. This comparison is shown in Figure \ref{fig:fig6}. In this case, our method successfully identifies and classifies the existing colors into the twelve categories created according to color theory. The method in \cite{mengibar2022learning} fails to create coherent categories, mixing the blues with the browns and the oranges with the greens. 
Additional examples of ABANICCO are available in the Supplementary Materials Section.

These experiments provide evidence regarding the scenarios in which ABANICCO can be of interest. For example, applications that deal with changing objects (such as video), cases where training is not possible or efficient (lack of data, for example), scenarios where edges are unclear or non-existent or when the number of classes is unknown, and especially cases where the user has no previous knowledge of the image or color theory. Furthermore, no extensive coding knowledge is required to run the proposed application. Moreover, these experiments further prove how easy it is for humans to misclassify colors due to context and perception.

In addition to these, we would like to call attention to how the simplicity of our method makes it computationally efficient, running fast even on low-end computers. On a mid-end computer (HP laptop with i5,2.4 Hz, and 16Gb RAM), it takes less than a second to analyze fully and provide a completely comprehensive description of all the shades present in the 380x570 pixels image in Figure \ref{fig:fig5}.

\newpage
\section{Conclusions}
\label{sec:conclusions}

In this paper, we have introduced ABANICCO: our AB Angular Illustrative Classification of Color, a method able to identify, segment, and classify all the different shades present in an image. Thus, it solves the pixel classification and color naming problems with a computationally efficient, geometrically-based solution. In addition, we have established a novel approach for modeling the semantics of color terms regardless of context and perception bias. The proposal has been successfully tested with the ISCC–NBS System of Color Designation and compared with the state of the art, yielding similar results without requiring manual annotations or training.

The main advantages of the proposed method are robustness and, most importantly, completely automatic and non-supervised character. ABANICCO does not require any human interaction, which means no training. This is especially important for computer vision due to the perception and context-driven errors humans inevitably make when classifying colors. It also means ABANICCO can be especially useful in scenarios where deep learning-based techniques are either unavailable or inefficient, as when labeled data is scarce and not enough to train supervised methods. Furthermore, as we have previously mentioned while analyzing Figure \ref{fig:fig5}, humans usually misclassify colors, which entails that supervised algorithms will be trained on biased data, not reaching the expected results.

The limitations of this pipeline are purely task-based, as it is meant for color segmentation and description,  not object segmentation. However, the results can be easily tweaked to fit the object segmentation once the user has seen the whole description of the colors present. Images with a lot of white may need this type of additional steps due to other colors reflecting on white objects.
We expect ABANICCO to be useful not only as a stand-alone application but also as an initial step in more complex tasks in Deep Learning, mainly training and labeling. We believe that our method represents a considerable improvement by providing automatic labels to reduce training time and improve results in a wide range of tasks such as the introduction of semantic information as organized color data \cite{bhowmick2022non,wheeler2021semantically}, sampling and registration \cite{saval20183d}, compression \cite{baig2017multiple}, image enhancement and dehazing \cite{jiang2017image}, and soft-labeling \cite{sucholutsky2020less}.

In future work, we would like to introduce the name of each color as additional modifiers to describe its pureness and lightness. This would help account for desaturated colors without figuring out their fully connected space \cite{jraissati2018delving}. Moreover, the code could be made to be even more efficient, successfully decreasing running time, which would make ABANICCO ideal for video processing in real-time.
All the data, code, resources, and additional materials used in our research are available in: \url{https://github.com/lauranicolass/ABANICCO}.


\section{Supplementary material}
\label{sec:Supplementary}

The easiest way to obtain the boundaries between the located color bases is to find each angle bisector. To avoid overlapping with the generated regions, those chromogen coordinate points with angles smaller than 5º were merged, assuming they would belong to the same region. This is shown in step 3 of Figure \ref{fig:fig2} 
We also tested the primary state-of-the-art classification methods against the bisector option. For this purpose, we used MATLAB’s classification learner. We created classes of 100 points laying over each of the proposed chromogen bases from the center to the limits of the reduced AB colorspace, with the bases extended to occupy the totality of the AB space. This was done to ensure that the classification methods would learn the regions entirely instead of leaving the edges undefined. With these points, we trained the following state-of-the-art classifiers using 5 cross-validation folds: Decision Trees (Fine, Medium and Coarse), Discriminant Analysis, Naïve Bayes, Support Vector Machines, Nearest Neighbor (Fine, Medium, Coarse, Cosine, Cubic, and Weighted), Kernel Approximations, Ensemble (Boosted Trees, Bagged Trees, Subspace Discriminant, Subspace KNN, and RUSBoosted Trees), and Neural Network (Narrow, Medium, Wide, Bilayered and Trilayered). For details on the specifics of how each model is trained, refer to Mathworks’ page on classification learner: \url{https://www.mathworks.com/help/stats/classificationlearner-app.html}.

Out of all the classifiers, the ones yielding the best results were Fine and Wide KNN, SVM, and Wide Neural Network classifiers. All these provided a validation accuracy larger than 0.98 and classified almost all the given points correctly. However, the resulting regions did not match the expected ones and failed to provide acceptable pixel classification when applied to actual images. Namely, they mix different colors in the same regions (as seen in comparative Figure \ref{fig:fig3}).  These approaches had good accuracy prediction but were unsuccessful when creating the expected color regions. This comparative analysis proved that the best regions were those obtained using the geometrical approach of the angle bisectors.


\section*{Acknowledgments}
This was partly supported by Ministerio de Ciencia, Innovacción y Universidades, Agencia Estatal de Investigación, under gran PID2019-109820RB, MCIN/AEI/10.13039/501100011033 co-financed by European Regional Development Fund (ERDF), "A way of making Europe" to A.M.B and L.N.S.
We would like to thank M Mengíbar-Rodríguez for her help in the results section.
\bibliographystyle{unsrt}  
\bibliography{references}

\end{document}